\documentclass{article}
\pdfoutput=1
\usepackage[final,nonatbib]{neurips_data_2023}

\usepackage[utf8]{inputenc} 
\usepackage[T1]{fontenc}    
\usepackage{hyperref}       
\usepackage{url}            
\usepackage{booktabs}       
\usepackage{amsfonts}       
\usepackage{nicefrac}       
\usepackage{microtype}      
\usepackage{xcolor}         
\usepackage{graphics}
\usepackage{graphicx}
\usepackage{tabularx}
\usepackage{subfig}

\title{Digital Typhoon: Long-term Satellite Image Dataset for the Spatio-Temporal Modeling of Tropical Cyclones}

\author{%
  Asanobu Kitamoto$^{1,2}$ \quad
  Jared Hwang$^{3,1}$ \quad
  Bastien Vuillod$^{4,1}$ \quad
  Lucas Gautier$^{5,1}$ \AND
  Yingtao Tian$^{6}$ \quad
  Tarin Clanuwat$^{6}$ \\ \\
  $^{1}$ National Institute of Informatics, Japan \\
  $^{2}$ Typhoon Science and Technology Research Center, Yokohama National University, Japan \\
  $^{3}$ University of Southern California, USA \\
  $^{4}$ Grenoble-INP, Ensimag, France \\
  $^{5}$ Université Clermont Auvergne, ISIMA, France \\
  $^{6}$ Google DeepMind\\
  }

\begin{document}

\maketitle

\begin{abstract}
  This paper presents the official release of the Digital Typhoon dataset, the longest typhoon satellite image dataset for 40+ years aimed at benchmarking machine learning models for long-term spatio-temporal data. To build the dataset, we developed a workflow to create an infrared typhoon-centered image for cropping using Lambert azimuthal equal-area projection referring to the best track data. We also address data quality issues such as inter-satellite calibration to create a homogeneous dataset. To take advantage of the dataset, we organized machine learning tasks by the types and targets of inference, with other tasks for meteorological analysis, societal impact, and climate change. The benchmarking results on the analysis, forecasting, and reanalysis for the intensity suggest that the dataset is challenging for recent deep learning models, due to many choices that affect the performance of various models. This dataset reduces the barrier for machine learning researchers to meet large-scale real-world events called tropical cyclones and develop machine learning models that may contribute to advancing scientific knowledge on tropical cyclones as well as solving societal and sustainability issues such as disaster reduction and climate change. The dataset is publicly available at  \url{http://agora.ex.nii.ac.jp/digital-typhoon/dataset/} and \url{https://github.com/kitamoto-lab/digital-typhoon/}.
\end{abstract}

\section{Introduction}

Tropical cyclones, also known as typhoons and hurricanes in certain regions, have been the critical target of research due to their substantial societal impact \cite{emanuel_100_2018}. To reduce the impact of tropical cyclones, the meteorological community, along with other earth science communities, has been developing both a theoretical and an empirical understanding of tropical cyclones through efforts such as advancing satellite remote sensing and atmospheric simulation models of higher spatial, temporal, and spectral resolutions for better analysis and forecasting.

Meteorologists have also developed an empirical method, known as the Dvorak technique \cite{dvorak_tropical_1975,velden_dvorak_2006}, to estimate the intensity of a tropical cyclone based on time-series observation data collected from worldwide ground sensor networks, meteorological satellites, and reconnaissance flights. This technique consists of a manual procedure to estimate tropical cyclone intensity based on the cloud patterns of satellite images and a temporal model for intensity change. The method was originally developed in the United States in the 1970s and later adopted by meteorological agencies worldwide to become the standard procedure. However, experts are aware of its heuristic and subjective nature, as it relies on empirical, rather than theoretical, human interpretation of observation data. Solutions to this problem include more objective and automated versions of the Dvorak technique\cite{olander_advanced_2007,olander_2021} and a citizen science project to take advantage of collective intelligence \cite{hennon_cyclone_2015}.

\begin{figure}[t]
  \centering
  \includegraphics[width=0.9\textwidth]{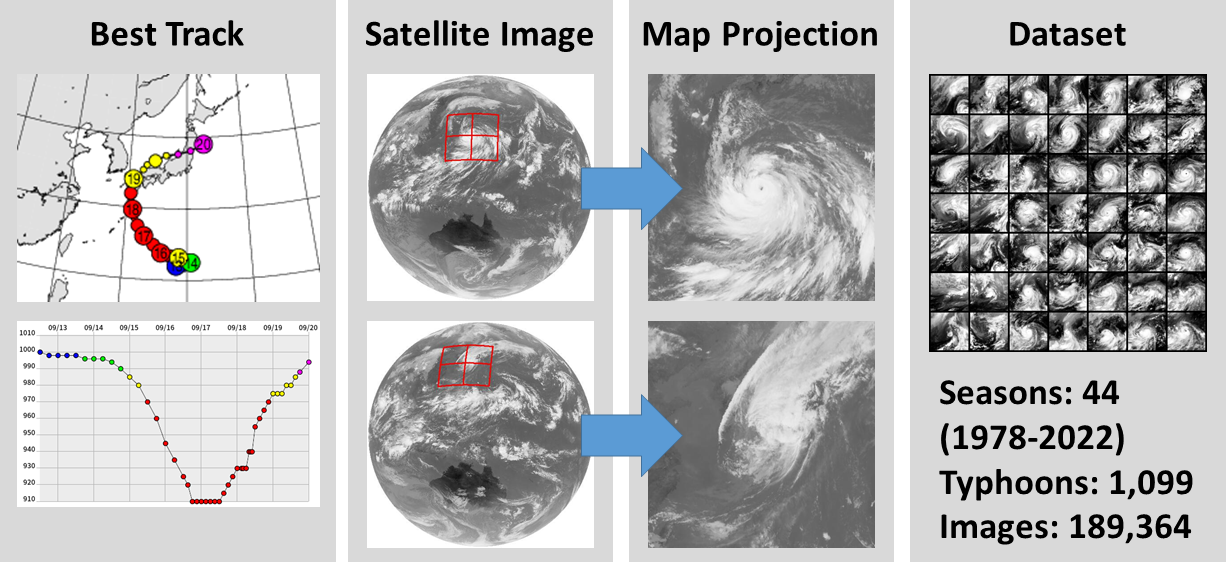}
   \caption{Overview of the Digital Typhoon dataset.}
  \label{fig:creation}
\end{figure}

It is clear that the Dvorak technique naturally fits into the machine learning framework by using images as input and intensity values as output. Hence there is a growing interest in both the machine learning community \cite{pradhan_tropical_2018,dawood_deep-phurie_2020,maskey_deepti_2020,lee_tropical_2020} and the meteorology community \cite{jaiswal_cyclone_2012,chen_estimating_2019} to take advantage of the big data of tropical cyclones for developing data-driven approaches. One of the authors, Asanobu Kitamoto, started the Digital Typhoon project in 1999 with the aim of applying machine learning to typhoon analysis and forecasting \cite{adinfo00,nii1}. The first step was to develop a homogeneous satellite image dataset for machine learning as in Figure~\ref{fig:creation}. The second step was to apply machine learning algorithms available at the time, such as SVM \cite{kitamoto_typhoon_2002}, Generative Topographic Mapping \cite{kitamoto_evolution_2002}, and content-based image retrieval\cite{kitamoto_spatio-temporal_2002}, which is later evolved into deep learning-based models for classification and regression tasks \cite{rodes-guirao_deep_2019,ci19}, combined with fisheye preprocessing \cite{higa_domain_2021}. The third step was to release the website "Digital Typhoon" in 2003 for browsing and searching datasets \cite{isde05}. The remaining problem was the lack of public datasets for machine learning. There have been attempts to download the dataset via scraping of the website (e.g. \cite{tian_tropical_2019}), but the dataset created in this way is of lower quality. 

Here we introduce the Digital Typhoon dataset, the {\em longest} typhoon satellite image dataset. This dataset reduces the burden of researchers to start machine learning on tropical cyclones without solid domain knowledge of meteorology and satellite remote sensing. We also illustrate the variety of tasks so that researchers can concentrate on building and evaluating machine-learning models.  

\section{Related Work}

\subsection{Track Datasets}
\label{sec:track}
The track data includes the 'annotation' of tropical cyclones, such as location, intensity, and wind circles, based on the interpretation of meteorological experts following the established procedure (e.g. Dvorak Technique). The best estimate, obtained from a retrospective analysis after collecting all the information from the start to the end of life, is called the best track dataset. 

The Digital Typhoon dataset targets the Western North Pacific basin, and the Japan Meteorological Agency (JMA) is designated as the regional center to maintain the best track dataset. 
Globally, the International Best Track Archive for Climate Stewardship (IBTrACS) \cite{knapp_international_2010} collects the best track from meteorological agencies worldwide and creates a comprehensive track dataset since 1842. 

IBTrACS shows an interesting variation of the best track; namely the location and intensity of the same tropical cyclone show discrepancies across meteorological agencies \cite{schreck_impact_2014}. This fact suggests that the interpretation of the observation data is not unique, or not the {\em ground truth} in a strict sense. Nonetheless, we regard the best track as the ground truth for most machine learning tasks, because it is the best estimate available. In a reanalysis task, however, we could critically evaluate the quality of the best track \cite{ito_analysis_2018}.

\subsection{Image Datasets}
\label{sec:image}
The image dataset has information about the spatial distribution of physical properties such as cloud patterns as grid data. 
The observation dataset \cite{knapp_scientific_2008,kossin_new_2007} is derived from sensor observation that measures the physical properties of the atmosphere, while the simulation dataset, both typhoon-related \cite{matsuoka_tropical_2023} and the global atmosphere \cite{rasp_weatherbench_2020,
ashkboos_ens-10_2022,cachay_climart_2021,racah_extreme_2017}, is generated as the representation of the atmosphere in a simulation model. Observation datasets and simulation datasets are linked through data assimilation, which is a statistical method to integrate observation datasets into a simulation model. 

The Digital Typhoon dataset is an observation dataset, and it offers a richer detail of tropical cyclones with higher temporal and spatial resolutions than the simulation dataset. In addition, data quality issues in the observation dataset, such as sensor noise, missing data, and long-term sensor calibration, are handled properly so that machine learning models are not significantly affected by those issues. 

\section{Digital Typhoon Dataset}

\begin{figure}[t]
   \centering
  \subfloat[Number of images for each grade]{\includegraphics[height=0.26\textwidth]{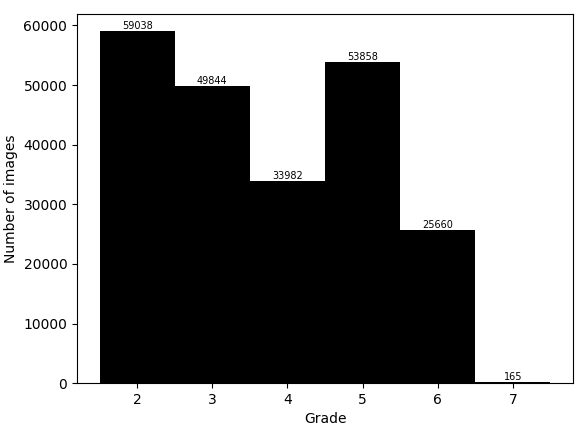}\label{fig:histo_grade}}
  \hfill
  \subfloat[Transition matrix for the grade]
  {\includegraphics[height=0.26\textwidth]{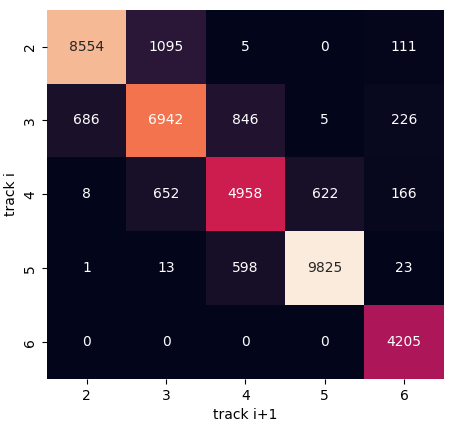}\label{fig:transition_matrix_grade}}
  \hfill
    \subfloat[Number of typhoons by the number of images]{\includegraphics[height=0.26\textwidth]{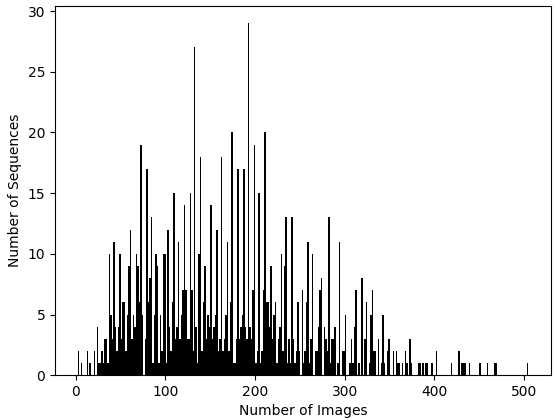}\label{fig:histo_sequence}}
    \caption{Visualization of statistics of the Digital Typhoon dataset.}
    \label{fig:statistics}
\end{figure}

\subsection{Dataset Overview}
\label{dataset-overview}

The Digital Typhoon dataset is created from the comprehensive satellite image archive of the Japanese geostationary satellite series, Himawari, from Himawari-1 to Himawari-9. Although those images are not copyrighted, some data are not accessible for free, and old satellite images have old formats for which open-source parsers are difficult to find. Hence we developed our own parsers for all generations of satellites, and the workflow to create typhoon-centered images by referring to the best track, as shown in Figure~\ref{fig:creation}. 

Using this workflow, we created the Digital Typhoon dataset by integrating metadata and images. The metadata contains hourly best-track data with additional information about the file name and each image's quality. The formatting of the best track data aligns with the original best track data sourced from the JMA. On the other hand, the images feature a 2D array of brightness temperatures around the typhoon's center, formatted in HDF5.  

As a result, the dataset comprises a total of 1,099 typhoons and 189,364 images. Figure~\ref{fig:statistics} visualizes some of the statistics of the dataset. It is a complete record of typhoons occurring in the Western North Pacific region (ranging from 100 to 180 degrees east of the northern hemisphere), from the 1978 season through the 2022 season, with missing typhoons in 1979 and 1980 due to the unavailability of satellite data.  The length of the dataset, spanning 44 typhoon seasons (years), is the {\em longest} typhoon image dataset. We call it the longest dataset because Japanese geostationary satellite images for typhoons before 1978 were lost forever, and our dataset went back to the oldest satellite image preserved. Hence it provides a unique opportunity to challenge long-term datasets.  

The Digital Typhoon dataset can reduce the burden of machine learning researchers to study tropical cyclones. First, it opens up access to tropical cyclone data processed from long-term satellite data. Second, it offers a homogeneous dataset created by the image processing workflow based on expertise in meteorology and satellite remote sensing. Third, massive computations to process hundreds of terabytes of original satellite data to create a machine-learning dataset are not necessary. A comprehensive explanation of the workflow for the creation of the dataset is provided in the Appendix.

The Digital Typhoon dataset is available at the official page \url{http://agora.ex.nii.ac.jp/digital-typhoon/dataset/} with an open data license, namely the Creative Commons Attribution 4.0 International (CC BY 4.0) License. 

\subsection{Comparison with the HURSAT Dataset}
\label{sec-hursat}

\begin{table}[t]
    \caption{Comparison between the Digital Typhoon dataset and the HURSAT dataset.}
    \label{tab:dt-hursat}
    \centering
    \begin{tabularx}{\textwidth}{lXX} 
    \toprule
               & Digital Typhoon dataset  &  HURSAT dataset \\ 
      \midrule
     Temporal coverage & 1978-2022 (present) & 1978-2015 \\
    Temporal resolution & one hour & three hours \\
     Target satellites & Himawari & SMS, GOES, Meteosat, Himawari, FY2 \\
     Spatial coverage & Western North Pacific basin 
     & All basins (Global) \\
     Spatial resolution & 5km & 8km \\
     Image coverage & 512$\times$512 pixels (1250km from the center) & 301$\times$301 pixels (1100km from the center) \\
      Spectral coverage & infrared (others on the Website) & visible, infrared, water vapor, near IR, split window \\
     Map projection & Azimuthal equal-area projection & Equirectangular projection \\
     Calibration & Recalibration & ISCCP \\
     Data format & HDF5 & NetCDF \\
     Best track & Japan Meteorological Agency & IBTrACS \\
     Dataset browsing & Digital Typhoon website & Download only \\
     \bottomrule
    \end{tabularx}
\end{table}

Among satellite image datasets of tropical cyclones, Hurricane Satellite Data (HURSAT) dataset \cite{knapp_scientific_2008,kossin_new_2007} from The National Oceanic and Atmospheric Administration (NOAA) is the most notable dataset in size and coverage. Table~\ref{tab:dt-hursat} provides a comparative summary of the Digital Typhoon and HURSAT datasets. There are distinct variations between the two as enumerated below.

\paragraph{Temporal coverage} The Digital Typhoon dataset is continually updated, and is the {\em longest} tropical cyclone image dataset worldwide. On the other hand, the HURSAT dataset stopped updating in 2015.

\paragraph{Temporal resolution} The Digital Typhoon dataset has a temporal resolution of one hour which is higher than the HURSAT dataset's three-hour resolution. A high-frequency change such as rapid intensification is more sensitive to temporal resolution. 

\paragraph{Spatial coverage} The Digital Typhoon dataset specifically targets the Western North Pacific basin, whereas the HURSAT dataset encompasses all basins.

\paragraph{Spatial resolution} The Digital Typhoon dataset possesses a spatial resolution of approximately 5km, superior to the HURSAT dataset's roughly 8km (0.07 degree). A small-scale structure such as the eye of a tropical cyclone is more sensitive to spatial resolution. 

\paragraph{Spectral coverage} The Digital Typhoon dataset incorporates the infrared (IR) channel, while the HURSAT dataset has more channels. 
It should be noted, however, that the Digital Typhoon {\em website} has the same spectral coverage, and the {\em dataset} can be easily extended to cover these channels.

\paragraph{Map projection} The Digital Typhoon dataset utilizes the Lambert azimuthal equal-area projection, maintaining the spherical shape of the tropical cyclone, while the HURSAT dataset employs the equirectangular (lat/long) grid, causing shape distortion in higher latitude or peripheral areas. Figure~\ref{fig:creation} shows an example of distortion when a typhoon is observed in the north. 

\paragraph{Dataset browsing} The Digital Typhoon {\em dataset} can be browsed via the Digital Typhoon {\em website}, which offers additional data. In contrast, the HURSAT dataset is solely available for download.

\subsection{Design Choices}

The dataset has several design choices, such as spectral coverage, spatial resolution, temporal resolution, and spatial coverage. In the following, we explain the reasons behind our choices.

\paragraph{Spectral coverage}
The current dataset includes only the Infrared channel (IR1) (wavelength of around $11\mu$m) but does not include any other channels available on the Digital Typhoon website. The following is the summary of the availability of each channel on the website. 
\begin{itemize}
\item IR1 (infrared): the data has been available since the beginning (1978).
\item VIS (visible): the data has been available since the beginning (1978), but images from early satellites were too noisy and not appropriate for a machine learning dataset. In addition, the visible channel is meaningful only during the daytime.
\item IR2 (infrared) and WV (water vapor): the data has been available since 1995 (Himawari-5).
\item NIR (near infrared) and other channels: the data has been available since 2005 (2nd generation) or 2015 (3rd generation).
\end{itemize}
As summarized, the IR1 is the only channel that is the longest and with fewer data quality issues, and this is the reason we included only the IR1 channel in our first version of the dataset. Future inclusion of multispectral data may offer additional tasks such as multispectral classification and regression. 

\paragraph{Spatial resolution}
The spatial resolution of about 5km per pixel reflects the spatial resolution of the IR1 channel for the first-generation satellites from Himawari-1 to Himawari-5. This resolution has improved to 4km for the second generation and 2km for the third generation. In spite of these progresses in technology, we chose a 5km resolution because it is the best choice to create a long-term homogeneous dataset. An interesting task in the future is to transfer a machine-learning model from long-term lower-resolution datasets to short-term higher-resolution datasets so that we can take advantage of recent technology for better forecasting.  

\paragraph{Temporal resolution}
The temporal resolution of one hour reflects the temporal resolution of one hour for some of the first-generation satellites after Himawari-3. From Himawari-1 to Himawari-2, the temporal resolution was more than one hour, or typically every three hours. For this reason, the data before 1987 has many missing data points as an hourly dataset. This resolution has improved to 30 minutes for the second generation and 10 minutes for the third generation. In spite of these progresses in technology, we chose one hour because it is a representative interval for many types of meteorological observations. 

\paragraph{Spatial coverage}
The current dataset only covers the Western North Pacific basin in the northern hemisphere (NH), but the Digital Typhoon website offers the same types of images for the southern hemisphere (SH) in the Australian basin using the best track from the Bureau of Meteorology, Australia. Here an interesting question is how a model trained in NH can be transferred to SH. From a meteorological point of view, tropical cyclones in various basins are considered the same meteorological phenomena, so theoretically, the dataset can be created similarly, and machine learning results are transferable. However, we also need to consider many details that may have an impact on the actual results, such as different quality of the best track data, and different sensor characteristics and calibration methods for different satellites. A future version of our datasets and benchmarks may address these issues. 

\section{Machine Learning Tasks}

The Digital Typhoon dataset serves two important roles. First, it offers a practical real-world dataset and tasks for the machine learning community to explore new models and solutions. Second, it provides a tool for meteorologists to apply data-driven approaches in studying tropical cyclones. The following is a summary of tasks in multiple dimensions. Other lists of tasks can be found in the review \cite{chen_machine_2020,wang_review_2022,reichstein_deep_2019}.

\subsection{Types of Inference}
\label{sec:inference}

\paragraph{Analysis} The task is to estimate current values using the current and past data. For instance, estimating the intensity of a typhoon falls into an analysis task, as it produces information about the typhoon's intensity using both current and past data. Supervised learning within this task can be further categorized into either a classification task or a regression task, contingent on whether the target variable is categorical or numerical. Additionally, unsupervised tasks can be designed for clustering or identifying typhoons with similar characteristics.

\paragraph{Forecasting} The task is to produce future predictions based on current and past data. The forecasts can be evaluated with the actual outcomes from the real event which become available over time. The forecasting task has a sub-task called nowcasting, aimed at making short-term forecasting spanning several hours using data-driven extrapolation. Note, however, that weather forecasting is theoretically constrained by the atmosphere's chaotic nature, which states that a minor difference in initial conditions can escalate over time. 

We call this task 'forecasting' instead of 'prediction' because prediction is ambiguous in machine learning. In meteorology, prediction is strictly used to mean future values, but in machine learning prediction could mean the output of a machine learning model without temporal dimension. To avoid confusion across disciplines, we use forecasting throughout the paper. 
 
\paragraph{Reanalysis} The task is to produce the best estimate given all obtainable data. This task is especially relevant to producing a uniform dataset spanning a long period of time, such as detecting trends in tropical cyclone activity to study the effects of climate change. As addressed in Section~\ref{sec:track}, the best track dataset may contain errors due to technological limitations or inconsistencies from different human experts. Machine learning can potentially aid in evaluating the quality of annotated data. 

\subsection{Targets of Inference}
\label{subsec-inference}

\paragraph{Intensity} The task makes inferences on the strength and size of a typhoon. The categorical grade is used to classify both the strength and type of a tropical cyclone. A classification task uses these grades as target variables. On the other hand, the intensity of tropical cyclones is measured numerically by central pressure and maximum sustained wind. An intensity regression task uses either pressure or wind as the target variable. In addition, the metadata includes the radius of the strong wind circle that represents the size of a tropical cyclone, so we can also design a regression task for size using the radius as the target variable.

\paragraph{Track} The task makes inferences on the geographical location of a typhoon. The cyclone's center, as estimated by human experts, is represented by latitude and longitude coordinates with a precision of 0.1 degrees. A regression task for predicting the typhoon's location uses these latitude and longitude coordinates as target variables.

\paragraph{Formation} The task makes inferences on the birth of a tropical cyclone, which typically occurs in tropical regions. Among the numerous cloud clusters actively evolving in tropical regions, determining which one will evolve into a tropical cyclone presents a challenging forecasting task, making it a target for machine learning applications \cite{chen_hybrid_2019}.

\paragraph{Transition} The task makes inferences on the transition from a tropical cyclone to an extra-tropical cyclone, which typically occurs in mid-latitude regions. Two types of cyclones are conceptually distinct from a meteorological perspective, but as a natural phenomenon, they are continuous. The data-driven modeling of a continuous transition process connecting two discrete concepts is a machine-learning task.

\subsection{Meteorological Analysis}

Machine learning can also be applied to analyze meteorological events on tropical cyclones, such as rapid intensification \cite{bhatia_recent_2019}, eyewall replacement \cite{hendricks_summary_2019}, and overshooting cloud tops \cite{kim_detection_2017}. These events may be linked with the forecasting of tropical cyclones, yet their underlying mechanisms are not entirely understood. Data-driven methodologies could potentially provide insights that contribute to the development of a novel theoretical framework for understanding these phenomena.

\subsection{Analysis for Societal Impact}

The Digital Typhoon dataset represents the atmospheric observation of a tropical cyclone, but its societal impacts are measured by different sources and modalities. For example, hazards are measured by heavy rainfall or strong winds, disasters are measured by landslides and flooding, and damages are measured by human casualties and financial loss. To construct a machine learning model to analyze and forecast the societal impact, real-world datasets from many sources should be integrated with meteorological datasets. This would enable a more comprehensive understanding of the full range of impacts arising from tropical cyclones.

\subsection{Analysis for Climate Change}

Understanding how a long-term tropical cyclone activity is impacted by climate change is a crucial topic in society \cite{landsea_can_2006,knutson_tropical_2010,kossin_trend_2013,rolnick_tackling_2022}. Technological and methodological evolution that occurred during the 40+ years lifespan introduces many types of biases in the dataset. While certain biases may be removed by sensor calibration, others are harder to detect such as annotation errors by human experts. The reanalysis of historical data and the creation of a homogeneous dataset can contribute to advancing our knowledge of the relationship between tropical cyclones and climate change.

\section{Benchmarks}
\label{sec-benchmarks}

\subsection{Overview of Benchmarks}

\paragraph{Task}
Machine learning tasks can be combined to create benchmarks for machine learning. We propose three benchmarks, 1) Analysis, 2) Forecasting, and 3) Reanalysis of the intensity of typhoons. The following summarizes some of the technical choices for benchmarking. 

\paragraph{Data splitting}
In meteorological time series, data are auto-correlated and one has to be careful how to split the data before starting to train a model \cite{schultz_can_2021}. At least, a random split for the image level must not be used to avoid overestimating the performance due to data leakage in the same typhoon sequence. Our assumption is that every sequence is independent, and we do not have to consider any leakage across sequences. So, as long as each sequence is treated as atomic when splitting the dataset, there is no limitation to using the entire dataset. Hence we apply random splits to the sequence level (split-by-sequence) or the season level (split-by-season). More complex splits can be designed, such as split by satellite generations (1978-2004, 2005-2014, 2015-2022). These designed splits are especially useful for the reanalysis task in Section~\ref{sec-reanalysis}.

\paragraph{Performance metric}

\begin{table}[t]
    \caption{The statistics of target values.}
    \label{tab:statistics-target}
    \centering
    \begin{tabular}{ccccc} 
    \toprule
    Target value & Range & Mean & Standard deviation  \\  \midrule
    Central pressure & 870-1018 (hPa) & 983.8 & 22.5
    \\  \midrule
    Maximum sustained wind & 35-140 (knots) & 59.2 & 19.8    \\  
           \bottomrule
        \end{tabular}
\end{table}

The following benchmarks evaluate the performance by the absolute error of target values because this is easier for domain experts to understand the result. However, for machine learning experts, the relative error of target values is more intuitive. Instead of showing absolute and relative errors for each benchmark, we summarize the statistics of target values so that relative errors can be roughly estimated. For example, the best result of $10.06\pm 0.09$ hPa RMSE in Table~\ref{tab:regression-pressure} is less than one standard deviation of error. 

\paragraph{Software and hardware}
To perform the benchmarks, we developed a Python-based software library \texttt{pyphoon2}, downloadable from \url{https://github.com/kitamoto-lab/digital-typhoon/}.
\texttt{pyphoon2} comes with a data loader and components to help build machine learning pipelines. All the experiments were performed on the internal cluster with 6 GPUs consisting of NVIDIA Quadro RTX 6000, NVIDIA Quadro RTX 8000, and NVIDIA Quadro RTX A6000.

\begin{table}[t]
    \caption{The result of the pressure regression task for two architectures and three types of input.}
    \label{tab:regression-pressure}
    \centering
    \begin{tabular}{ccccc} 
    \toprule
    \textbf{RMSE (hPa)} & Full ($512\times512$) & Resized ($224\times 224$) & Cropped ($224\times 224$)
    \\  \midrule
    ResNet18 & 10.51 ($\pm$0.11) & 10.47($\pm$0.20) & \textbf{10.06} ($\pm$0.09)
    \\  \midrule
    ResNet50 & 11.12 ($\pm$0.41) & 11.63 ($\pm$0.35) & 10.09 ($\pm$0.04)
    \\  
           \bottomrule
        \end{tabular}

\end{table}

\begin{table}[t]
    \caption{The result of the wind regression task for two architectures and three types of input.}
    \label{tab:regression-wind}
    \centering
    \begin{tabular}{ccccc} 
    \toprule
    \textbf{RMSE (kt)} & Full ($512\times512$) & Resized ($224\times 224$) & Cropped ($224\times 224$)
    \\  \midrule
    ResNet18 & 10.21 ($\pm$0.19) & 10.09 ($\pm$0.08) & \textbf{9.25} ($\pm$0.25)
    \\  \midrule
    ResNet50 & 10.05 ($\pm$0.26) & 10.21 ($\pm$0.14) & 9.13 ($\pm$0.11)
    \\ 
               \bottomrule
        \end{tabular}

\end{table}

\begin{figure}[t]
   \centering
  \subfloat[Plots for wind regression by ResNet18 for cropped images.]{\includegraphics[width=0.45\textwidth]{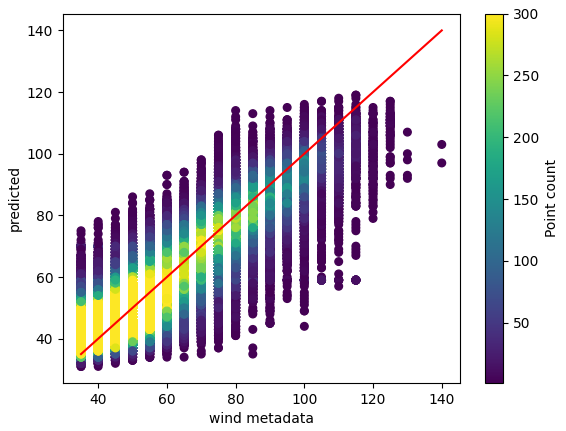}\label{fig:resnet_pred_wind_cropped}}
  \hfill
  \subfloat[Plots for pressure regression by ResNet18 for cropped images.]{\includegraphics[width=0.46\textwidth]{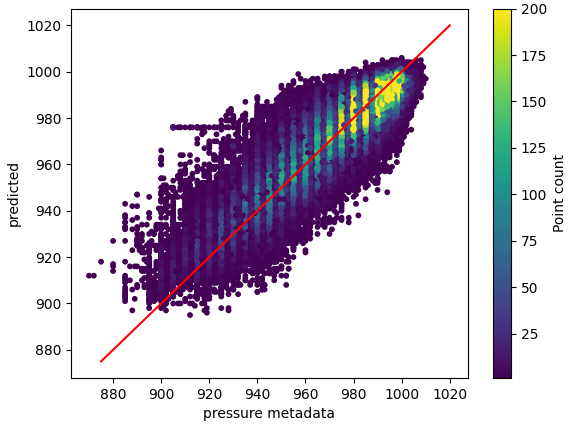}\label{fig:resnet_pred_pressure_cropped}}\\
\caption{Prediction plots for pressure and wind regressions.}
\label{fig:regression}
  \end{figure}

\subsection{Analysis for the Intensity}
\label{sec-analysis}

We propose classification tasks, which take an image as input and estimate grade as output, and regression tasks, which take an image as input and estimate a pressure or wind value as output.
In the JMA best track, grades 3, 4, and 5 denote a tropical cyclone, among which grade 5 is the most intense according to the maximum sustained wind. Grade 2 signifies a tropical depression, a type of cyclone weaker than a tropical cyclone. Moreover, grade 6 corresponds to an extra-tropical cyclone, a type of cyclone having a different structure from a tropical cyclone.
Central pressure in hectopascal (hPa) is recorded for all grades, while the maximum sustained wind in knot (kt) is recorded only for grades 3, 4, and 5. In the following, we describe the result of the regression task, and the result of the classification task is described in the appendix.

We explored three types of comparisons. First, we compared three architectures, namely VGG \cite{pytorchvgg}, ResNet \cite{pytorchresnet} and Vision Transformer \cite{pytorchtransformer}. Second, we compared models trained on 1) full-resolution images ($512\times512$), 2) resized images ($224\times224$), and 3) cropped images ($224\times224$). In 2), the full region of the image is resized, while in 3), the central region of the image is cropped without resizing. The latter is inspired by the Dvorak technique, which focuses on many relevant image features found around the typhoon center. Third, we compared two target values, namely pressure, and wind. 

We used the TorchVision \cite{torchvision2016} ResNet18 and ResNet50 models with a learning rate (LR) of $10^{-4}$, batch size of 16, and for 50 epochs. An 80/20 train/test split by sequence was used. The ResNet18 and ResNet50 models were trained five and two times respectively. To evaluate, we measured the root mean square error (RMSE) of the prediction from ground truth and their standard deviations ($\pm$ std). 

Table~\ref{tab:regression-pressure} and Table~\ref{tab:regression-wind} summarize the results. Firstly, ResNet50 yielded similar results to ResNet18. Secondly, cropping the images around the typhoon center yielded a lower RMSE than other choices, indicating that cropping is better than resizing in preserving features around the typhoon center, or removing non-relevant features far from the center. Training a model on the full images did not perform well due to their larger number of pixels. Furthermore, Figure~\ref{fig:regression} illustrates that regression performs better for weaker typhoons, but worse for stronger typhoons. 

\subsection{Forecasting for the Intensity}

\begin{figure}[t]
    \centering
    \includegraphics[width=0.85\textwidth]{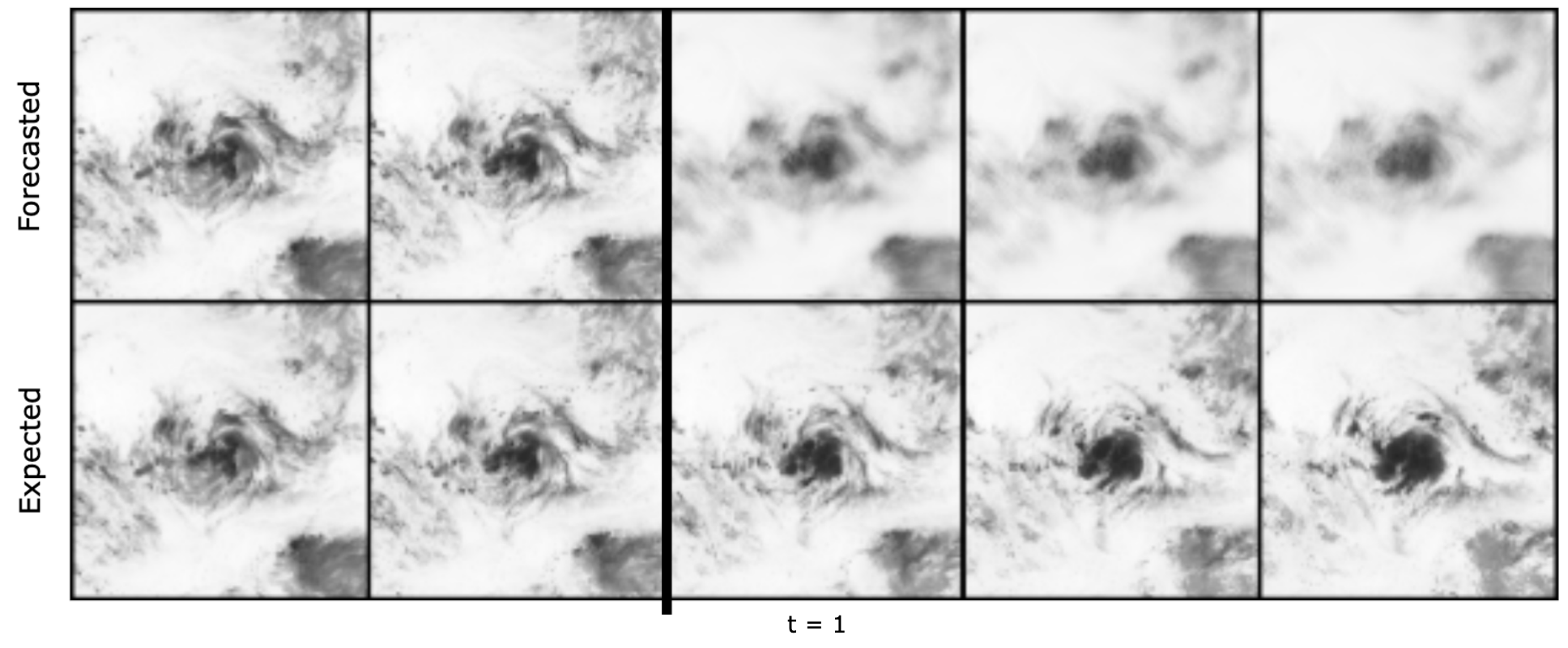}
    \caption[width=0.8\textwidth]{Results of image forecasting by ConvLSTM.}
    \label{fig:forecasting_img}
\end{figure}

\begin{table}[t]
    \caption{Results of pressure forecasting for 12-hours by ResNet18 (values in hPa).}
    \label{tab:forecast-table}
    \centering
    \begin{tabular}{cccccc} 
    \toprule
     $t$ & 1 & 2 & 3 & 6 & 12
     \\  \midrule
    \bf{RMSE} & $10.24 \pm 0.73$ & $10.52 \pm 0.79$ & $11.00 \pm 0.87$ & $12.10 \pm 0.85$ & $14.69 \pm 0.91$
    \\
       \bottomrule
    \end{tabular}
\end{table}

Our previous work used Recurrent Neural Network (RNN) to forecast the pressure directly from images and showed comparable performance with SHIPS \cite{ci19}. In this paper, we chose another approach using a convolutional LSTM \cite{shi2015convolutional} to predict the next $n$ image frames of a typhoon given the previous 12 image frames and analyze the pressure from the predicted image. We adapted an implementation \cite{convlstmimp}, and used a 3-layer ConvLSTM with 128 hidden dimensions. Due to resource limitations, we used $128\times 128$ downsampled images from only the first 24 hours of a given typhoon.

To forecast $n$ hours into the future (starting at $t=1$), the 12 preceding frames ($t=[-11, 0]$) are passed into the model, which outputs a single image serving as its forecast at $t=1$. Then, images from $t=[-10, 1]$, including the predicted image, are passed back into the model to get the prediction for $t=2$. This process is repeated $n$ times to forecast $n$ hours into the future. As a result, Figure~\ref{fig:forecasting_img} shows that the first predicted frame is perceptively blurred and rapidly deteriorates as $t$ advances.

We then trained a ResNet18 model on predicted images, as well as the first 24 images of every typhoon, to predict the pressure given a $128\times 128$ image. As a result, Table \ref{tab:forecast-table} shows that the model produces a larger RMSE and error as $t$ advances due to the blur of predicted images. A future adaptation may be to train both the ConvLSTM and ResNet in a black box, such that the loss is minimized by image reproduction and pressure prediction.

Both models were trained on the same 80/20 train/test split by sequence. The ConvLSTM was trained once for 230 epochs with a starting LR of $10^{-4}$, and used a CosineAnnealing scheduler \cite{loshchilov2017sgdr} with 100 steps. The ResNet model used a modified first convolutional layer with a kernel and stride size of (2, 2) and (1, 1). It was trained five times with an LR of $10^{-5}$ for 34 epochs. These hyperparameters were chosen as they produced more consistent results given the smaller image sizes. 

\subsection{Reanalysis for the Intensity}
\label{sec-reanalysis}
The goal of this paper is to create a homogeneous long-term dataset, and the purpose of the reanalysis task is to identify biases and inconsistencies in the dataset due to factors such as technological evolution arising from satellite sensors, or methodological evolution arising from the improvement of the Dvorak technique to annotate tropical cyclones. One approach to this challenge is to design special data splits to analyze historical factors. 

Our previous work studied this task by training the model using recent data and testing on past data to analyze the trend of model performance, indicating that old satellite data may have different characteristics \cite{ci19}. In this paper, we split the dataset into three buckets by satellite generations, namely the first generation (1978-2004), the second generation (2005-2014), and the third generation (2015-2022), and train and test a ResNet18 model for the regression task. 

Input images were resized to $224\times224$ in the same way as the analysis task. 208 sequences (the size of the smallest generation) were then randomly sampled from each generation five times and split into 80/20 train/test sets. A ResNet18 model was trained on each bucket, each for $101$ epochs with a batch size of $16$ and a learning rate of $10^{-4}$; these parameters were chosen for their consistent results. Each of the three models was then tested on the test set of each bucket, such that a model trained on the first bucket was tested on the first, second, and third buckets. 

\begin{table}[t]
    \caption{The results of regression tasks for each satellite generation (values in hPa).}  
     \label{tab:reanalysis-table}
    \centering
    \begin{tabular}{ccccc} 
    \toprule
    \textbf{RMSE} & Train the First & Train the Second & Train the Third
    \\  \midrule
    Test the First & 10.04 ($\pm$0.17) & 9.92 ($\pm$0.09) & 10.03 ($\pm$0.10)
    \\  \midrule
    Test the Second & 12.80 ($\pm$0.19) & 11.05 ($\pm$0.10) & 11.17 ($\pm$0.10)
    \\  \midrule
    Test the Third & 10.34 ($\pm$0.17) & 10.03 ($\pm$0.08) & 9.94 ($\pm$0.16)
    \\     \bottomrule
        \end{tabular}
\end{table}

Table~\ref{tab:reanalysis-table} shows that all three models performed roughly similarly on all three buckets, and no dataset bias was immediately reflected in the quality of the models. An expansion on the reanalysis task experiment we performed is described in the Appendix.

\subsection{Comparison with Other Approaches}
Machine learning is not the only approach for data-driven analysis and forecasting of tropical cyclones. For the analysis of intensity, the Dvorak technique has been the most popular method among meteorologists. In addition, for the forecasting of intensity, computational approaches represent a typhoon in a simulation model and compute the future based on the theory of the atmosphere. This approach, however, has limitations due to spatial and temporal resolutions, and intensity forecasting is still considered a difficult challenge. Instead, meteorologists have developed empirical methods, such as SHIPS \cite{fitzpatrick_understanding_1997,yamaguchi_tropical_2018} with linear regression on hand-crafted meteorological features, or a similar approach using XGBoost \cite{chan1002-2041}. 
This paper focused on the comparison of machine learning models, but the real challenge for domain experts is comparing not only machine learning approaches but also computational, empirical, or manual approaches in the context of real-world solutions for tropical cyclones, such as disaster reduction. This paper is a starting point for this grand challenge. 

\section{Conclusion}

We have introduced the Digital Typhoon dataset for machine learning and meteorology communities to promote data-driven research on tropical cyclones. Our dataset offers a unique opportunity to benchmark various types of machine learning models, especially spatio-temporal models for long-term time-series images. A solution is not only valuable for machine learning benchmarking but also has the potential to contribute to advancing scientific knowledge on tropical cyclones as well as solving societal and sustainability issues such as disaster reduction and climate change. 

\begin{ack}
Three of the authors, Jared Hwang, Bastien Vuillod, and Lucas Gautier, have been supported by the international internship program of the National Institute of Informatics.
\end{ack}

\newpage
\bibliographystyle{plain}
\bibliography{neurips_data_2023}

\begin{thebibliography}{10}

\bibitem{ashkboos_ens-10_2022}
Saleh Ashkboos, Langwen Huang, Nikoli Dryden, Tal Ben-Nun, Peter~Dominik Dueben, Lukas Gianinazzi, Luca~Nicola Kummer, and Torsten Hoefler.
\newblock {ENS}-10: {A} {Dataset} {For} {Post}-{Processing} {Ensemble} {Weather} {Forecasts}.
\newblock In {\em Thirty-sixth Conference on Neural Information Processing Systems Datasets and Benchmarks Track}, September 2022.

\bibitem{bhatia_recent_2019}
Kieran~T. Bhatia, Gabriel~A. Vecchi, Thomas~R. Knutson, Hiroyuki Murakami, James Kossin, Keith~W. Dixon, and Carolyn~E. Whitlock.
\newblock Recent increases in tropical cyclone intensification rates.
\newblock {\em Nature Communications}, 10(1):635, February 2019.
\newblock Number: 1 Publisher: Nature Publishing Group.

\bibitem{cachay_climart_2021}
Salva~Rühling Cachay, Venkatesh Ramesh, Jason N.~S. Cole, Howard Barker, and David Rolnick.
\newblock {ClimART}: {A} {Benchmark} {Dataset} for {Emulating} {Atmospheric} {Radiative} {Transfer} in {Weather} and {Climate} {Models}.
\newblock In {\em 35th {Conference} on {Neural} {Information} {Processing} {Systems} {(NeurIPS 2021)} {Track} on {Datasets} and {Benchmarks}}, October 2021.

\bibitem{chan1002-2041}
Ming Hei~Kenneth Chan, Wai~Kin Wong, and Kin~Chung Au-Yeung.
\newblock Machine learning in calibrating tropical cyclone intensity forecast of ecmwf eps.
\newblock {\em Meteorological Applications}, 28(6):e2041, 2021.

\bibitem{chen_estimating_2019}
Buo-Fu Chen, Boyo Chen, Hsuan-Tien Lin, and Russell~L. Elsberry.
\newblock Estimating {Tropical} {Cyclone} {Intensity} by {Satellite} {Imagery} {Utilizing} {Convolutional} {Neural} {Networks}.
\newblock {\em Weather and Forecasting}, 34(2):447--465, April 2019.
\newblock Publisher: American Meteorological Society Section: Weather and Forecasting.

\bibitem{chen_hybrid_2019}
Rui Chen, Xiang Wang, Weimin Zhang, Xiaoyu Zhu, Aiping Li, and Chao Yang.
\newblock A hybrid {CNN}-{LSTM} model for typhoon formation forecasting.
\newblock {\em GeoInformatica}, 23(3):375--396, July 2019.

\bibitem{chen_machine_2020}
Rui Chen, Weimin Zhang, and Xiang Wang.
\newblock Machine {Learning} in {Tropical} {Cyclone} {Forecast} {Modeling}: {A} {Review}.
\newblock {\em Atmosphere}, 11(7):676, July 2020.
\newblock Number: 7 Publisher: Multidisciplinary Digital Publishing Institute.

\bibitem{dawood_deep-phurie_2020}
Muhammad Dawood, Amina Asif, and Fayyaz ul Amir~Afsar Minhas.
\newblock Deep-{PHURIE}: deep learning based hurricane intensity estimation from infrared satellite imagery.
\newblock {\em Neural Computing and Applications}, 32(13):9009--9017, July 2020.

\bibitem{pytorchtransformer}
Alexey Dosovitskiy, Lucas Beyer, Alexander Kolesnikov, Dirk Weissenborn, Xiaohua Zhai, Thomas Unterthiner, Mostafa Dehghani, Matthias Minderer, Georg Heigold, Sylvain Gelly, Jakob Uszkoreit, and Neil Houlsby.
\newblock An image is worth 16x16 words: Transformers for image recognition at scale, 2021.

\bibitem{dvorak_tropical_1975}
Vernon~F. Dvorak.
\newblock Tropical {Cyclone} {Intensity} {Analysis} and {Forecasting} from {Satellite} {Imagery}.
\newblock {\em Monthly Weather Review}, 103(5):420--430, May 1975.

\bibitem{emanuel_100_2018}
Kerry Emanuel.
\newblock 100 {Years} of {Progress} in {Tropical} {Cyclone} {Research}.
\newblock {\em Meteorological Monographs}, 59:15.1 -- 15.68, 2018.
\newblock Place: Boston MA, USA Publisher: American Meteorological Society.

\bibitem{fitzpatrick_understanding_1997}
Patrick~J. Fitzpatrick.
\newblock Understanding and {Forecasting} {Tropical} {Cyclone} {Intensity} {Change} with the {Typhoon} {Intensity} {Prediction} {Scheme} ({TIPS}).
\newblock {\em Weather and Forecasting}, 12(4):826--846, December 1997.
\newblock Publisher: American Meteorological Society Section: Weather and Forecasting.

\bibitem{pytorchresnet}
Kaiming He, Xiangyu Zhang, Shaoqing Ren, and Jian Sun.
\newblock Deep residual learning for image recognition, 2015.

\bibitem{hendricks_summary_2019}
Eric~A. Hendricks, Scott~A. Braun, Jonathan~L. Vigh, and Joseph~B. Courtney.
\newblock A summary of research advances on tropical cyclone intensity change from 2014-2018.
\newblock {\em Tropical Cyclone Research and Review}, 8(4):219--225, December 2019.

\bibitem{hennon_cyclone_2015}
Christopher~C. Hennon, Kenneth~R. Knapp, Carl~J. Schreck, Scott~E. Stevens, James~P. Kossin, Peter~W. Thorne, Paula~A. Hennon, Michael~C. Kruk, Jared Rennie, Jean-Maurice Gadéa, Maximilian Striegl, and Ian Carley.
\newblock Cyclone {Center}: {Can} {Citizen} {Scientists} {Improve} {Tropical} {Cyclone} {Intensity} {Records}?
\newblock {\em Bulletin of the American Meteorological Society}, 96(4):591 -- 607, 2015.
\newblock Place: Boston MA, USA Publisher: American Meteorological Society.

\bibitem{higa_domain_2021}
Maiki Higa, Shinya Tanahara, Yoshitaka Adachi, Natsumi Ishiki, Shin Nakama, Hiroyuki Yamada, Kosuke Ito, Asanobu Kitamoto, and Ryota Miyata.
\newblock Domain knowledge integration into deep learning for typhoon intensity classification.
\newblock {\em Scientific Reports}, 11(1):12972, June 2021.
\newblock Number: 1 Publisher: Nature Publishing Group.

\bibitem{ito_analysis_2018}
Kosuke Ito, Hiroyuki Yamada, Munehiko Yamaguchi, Tetsuo Nakazawa, Norio Nagahama, Kensaku Shimizu, Tadayasu Ohigashi, Taro Shinoda, and Kazuhisa Tsuboki.
\newblock Analysis and {Forecast} {Using} {Dropsonde} {Data} from the {Inner}-{Core} {Region} of {Tropical} {Cyclone} {Lan} (2017) {Obtained} during the {First} {Aircraft} {Missions} of {T}-{PARCII}.
\newblock {\em Sola}, 14:105--110, 2018.

\bibitem{jaiswal_cyclone_2012}
Neeru Jaiswal, C.~M. Kishtawal, and P.~K. Pal.
\newblock Cyclone intensity estimation using similarity of satellite {IR} images based on histogram matching approach.
\newblock {\em Atmospheric Research}, 118:215--221, November 2012.

\bibitem{john_methods_2019}
Viju~O. John, Tasuku Tabata, Frank Rüthrich, Rob Roebeling, Tim Hewison, Reto Stöckli, and Jörg Schulz.
\newblock On the {Methods} for {Recalibrating} {Geostationary} {Longwave} {Channels} {Using} {Polar} {Orbiting} {Infrared} {Sounders}.
\newblock {\em Remote Sensing}, 11(10):1171, January 2019.
\newblock Number: 10 Publisher: Multidisciplinary Digital Publishing Institute.

\bibitem{kim_detection_2017}
Miae Kim, Jungho Im, Haemi Park, Seonyoung Park, Myong-In Lee, and Myoung-Hwan Ahn.
\newblock Detection of {Tropical} {Overshooting} {Cloud} {Tops} {Using} {Himawari}-8 {Imagery}.
\newblock {\em Remote Sensing}, 9(7):685, July 2017.
\newblock Number: 7 Publisher: Multidisciplinary Digital Publishing Institute.

\bibitem{adinfo00}
Asanobu KITAMOTO.
\newblock The development of typhoon image database with content-based search.
\newblock In {\em Proceedings of the 1st International Symposium on Advanced Informatics (AdInfo)}, pages 163--170, 3 2000.

\bibitem{kitamoto_evolution_2002}
Asanobu Kitamoto.
\newblock Evolution {Map}: {Modeling} {State} {Transition} of {Typhoon} {Image} {Sequences} by {Spatio}-{Temporal} {Clustering}.
\newblock In Steffen Lange, Ken Satoh, and Carl~H. Smith, editors, {\em Discovery {Science}}, Lecture {Notes} in {Computer} {Science}, pages 283--290, Berlin, Heidelberg, 2002. Springer.

\bibitem{kitamoto_spatio-temporal_2002}
Asanobu Kitamoto.
\newblock Spatio-{Temporal} {Data} {Mining} for {Typhoon} {Image} {Collection}.
\newblock {\em Journal of Intelligent Information Systems}, 19(1):25--41, July 2002.

\bibitem{kitamoto_typhoon_2002}
Asanobu Kitamoto.
\newblock Typhoon {Analysis} and {Data} {Mining} with {Kernel} {Methods}.
\newblock In Seong-Whan Lee and Alessandro Verri, editors, {\em Pattern {Recognition} with {Support} {Vector} {Machines}}, Lecture {Notes} in {Computer} {Science}, pages 237--249, Berlin, Heidelberg, 2002. Springer.

\bibitem{isde05}
Asanobu KITAMOTO.
\newblock Digital typhoon: Near real-time aggregation, recombination and delivery of typhoon-related information.
\newblock In {\em Proceedings of the 4th International Symposium on Digital Earth (ISDE)}, page 16 pages, 3 2005.

\bibitem{nii1}
Asanobu KITAMOTO and Kinji ONO.
\newblock The construction of typhoon image collection and its application to typhoon analysis.
\newblock {\em NII Journal}, 1:7--22, 12 2000.
\newblock (in Japanese).

\bibitem{knapp_scientific_2008}
Kenneth~R. Knapp.
\newblock Scientific data stewardship of international satellite cloud climatology project {B1} global geostationary observations.
\newblock {\em Journal of Applied Remote Sensing}, 2(1):023548, November 2008.

\bibitem{knapp_international_2010}
Kenneth~R. Knapp, Michael~C. Kruk, David~H. Levinson, Howard~J. Diamond, and Charles~J. Neumann.
\newblock The {International} {Best} {Track} {Archive} for {Climate} {Stewardship} ({IBTrACS}): {Unifying} {Tropical} {Cyclone} {Data}.
\newblock {\em Bulletin of the American Meteorological Society}, 91(3):363--376, March 2010.

\bibitem{knutson_tropical_2010}
Thomas~R. Knutson, John~L. McBride, Johnny Chan, Kerry Emanuel, Greg Holland, Chris Landsea, Isaac Held, James~P. Kossin, A.~K. Srivastava, and Masato Sugi.
\newblock Tropical cyclones and climate change.
\newblock {\em Nature Geoscience}, 3(3):157--163, March 2010.
\newblock Number: 3 Publisher: Nature Publishing Group.

\bibitem{kossin_new_2007}
James~P. Kossin.
\newblock New global tropical cyclone data set from {ISCCP} {B1} geostationary satellite observations.
\newblock {\em Journal of Applied Remote Sensing}, 1(1):013505, February 2007.

\bibitem{kossin_trend_2013}
James~P. Kossin, Timothy~L. Olander, and Kenneth~R. Knapp.
\newblock Trend {Analysis} with a {New} {Global} {Record} of {Tropical} {Cyclone} {Intensity}.
\newblock {\em Journal of Climate}, 26(24):9960--9976, December 2013.
\newblock Publisher: American Meteorological Society Section: Journal of Climate.

\bibitem{landsea_can_2006}
Christopher~W. Landsea, Bruce~A. Harper, Karl Hoarau, and John~A. Knaff.
\newblock Can {We} {Detect} {Trends} in {Extreme} {Tropical} {Cyclones}?
\newblock {\em Science}, 313(5786):452--454, July 2006.
\newblock Publisher: American Association for the Advancement of Science.

\bibitem{lee_tropical_2020}
Juhyun Lee, Jungho Im, Dong-Hyun Cha, Haemi Park, and Seongmun Sim.
\newblock Tropical {Cyclone} {Intensity} {Estimation} {Using} {Multi}-{Dimensional} {Convolutional} {Neural} {Networks} from {Geostationary} {Satellite} {Data}.
\newblock {\em Remote Sensing}, 12(1):108, January 2020.
\newblock Number: 1 Publisher: Multidisciplinary Digital Publishing Institute.

\bibitem{loshchilov2017sgdr}
Ilya Loshchilov and Frank Hutter.
\newblock Sgdr: Stochastic gradient descent with warm restarts, 2017.

\bibitem{torchvision2016}
TorchVision maintainers and contributors.
\newblock Torchvision: Pytorch's computer vision library.
\newblock \url{https://github.com/pytorch/vision}, 2016.

\bibitem{maskey_deepti_2020}
Manil Maskey, Rahul Ramachandran, Muthukumaran Ramasubramanian, Iksha Gurung, Brian Freitag, Aaron Kaulfus, Drew Bollinger, Daniel~J. Cecil, and Jeffrey Miller.
\newblock Deepti: {Deep}-{Learning}-{Based} {Tropical} {Cyclone} {Intensity} {Estimation} {System}.
\newblock {\em IEEE Journal of Selected Topics in Applied Earth Observations and Remote Sensing}, 13:4271--4281, 2020.
\newblock Conference Name: IEEE Journal of Selected Topics in Applied Earth Observations and Remote Sensing.

\bibitem{matsuoka_tropical_2023}
Daisuke Matsuoka, Chihiro Kodama, Yohei Yamada, and Masuo Nakano.
\newblock Tropical cyclone dataset for a high-resolution global nonhydrostatic atmospheric simulation.
\newblock {\em Data in Brief}, 48:109135, June 2023.

\bibitem{olander_2021}
Timothy Olander, Anthony Wimmers, Christopher Velden, and James~P. Kossin.
\newblock Investigation of machine learning using satellite-based advanced dvorak technique analysis parameters to estimate tropical cyclone intensity.
\newblock {\em Weather and Forecasting}, 36(6):2161 -- 2186, 2021.

\bibitem{olander_advanced_2007}
Timothy~L. Olander and Christopher~S. Velden.
\newblock The {Advanced} {Dvorak} {Technique}: {Continued} {Development} of an {Objective} {Scheme} to {Estimate} {Tropical} {Cyclone} {Intensity} {Using} {Geostationary} {Infrared} {Satellite} {Imagery}.
\newblock {\em Weather and Forecasting}, 22(2):287--298, April 2007.
\newblock Publisher: American Meteorological Society Section: Weather and Forecasting.

\bibitem{convlstmimp}
Rohit Panda.
\newblock Video frame prediction using convlstm network in pytorch, 6 2021.

\bibitem{ci19}
Clément Playout and Asanobu KITAMOTO.
\newblock Latent space representation and rnn for image-based typhoon intensity analysis and prediction.
\newblock In {\em The 9th International Workshop on Climate Informatics (CI2019)}, pages 47--52, 10 2019.

\bibitem{pradhan_tropical_2018}
Ritesh Pradhan, Ramazan~S. Aygun, Manil Maskey, Rahul Ramachandran, and Daniel~J. Cecil.
\newblock Tropical {Cyclone} {Intensity} {Estimation} {Using} a {Deep} {Convolutional} {Neural} {Network}.
\newblock {\em IEEE Transactions on Image Processing}, 27(2):692--702, February 2018.

\bibitem{racah_extreme_2017}
Evan Racah, Christopher Beckham, Tegan Maharaj, Samira~Ebrahimi Kahou, Prabhat, and Christopher Pal.
\newblock Extreme weather: a large-scale climate dataset for semi-supervised detection, localization, and understanding of extreme weather events.
\newblock In {\em Proceedings of the 31st {International} {Conference} on {Neural} {Information} {Processing} {Systems}}, {NIPS}'17, pages 3405--3416, Red Hook, NY, USA, December 2017. Curran Associates Inc.

\bibitem{rasp_weatherbench_2020}
Stephan Rasp, Peter~D. Dueben, Sebastian Scher, Jonathan~A. Weyn, Soukayna Mouatadid, and Nils Thuerey.
\newblock {WeatherBench}: {A} {Benchmark} {Data} {Set} for {Data}-{Driven} {Weather} {Forecasting}.
\newblock {\em Journal of Advances in Modeling Earth Systems}, 12(11):e2020MS002203, 2020.
\newblock \_eprint: https://onlinelibrary.wiley.com/doi/pdf/10.1029/2020MS002203.

\bibitem{reichstein_deep_2019}
Markus Reichstein, Gustau Camps-Valls, Bjorn Stevens, Martin Jung, Joachim Denzler, Nuno Carvalhais, and {Prabhat}.
\newblock Deep learning and process understanding for data-driven {Earth} system science.
\newblock {\em Nature}, 566(7743):195--204, February 2019.

\bibitem{rodes-guirao_deep_2019}
Lucas Rodés-Guirao.
\newblock Deep {Learning} for {Digital} {Typhoon} : {Exploring} a typhoon satellite image dataset using deep learning, 2019.

\bibitem{rolnick_tackling_2022}
David Rolnick, Priya~L. Donti, Lynn~H. Kaack, Kelly Kochanski, Alexandre Lacoste, Kris Sankaran, Andrew~Slavin Ross, Nikola Milojevic-Dupont, Natasha Jaques, Anna Waldman-Brown, Alexandra~Sasha Luccioni, Tegan Maharaj, Evan~D. Sherwin, S.~Karthik Mukkavilli, Konrad~P. Kording, Carla~P. Gomes, Andrew~Y. Ng, Demis Hassabis, John~C. Platt, Felix Creutzig, Jennifer Chayes, and Yoshua Bengio.
\newblock Tackling {Climate} {Change} with {Machine} {Learning}.
\newblock {\em ACM Computing Surveys}, 55(2):42:1--42:96, February 2022.

\bibitem{schreck_impact_2014}
Carl~J. Schreck, Kenneth~R. Knapp, and James~P. Kossin.
\newblock The {Impact} of {Best} {Track} {Discrepancies} on {Global} {Tropical} {Cyclone} {Climatologies} using {IBTrACS}.
\newblock {\em Monthly Weather Review}, 142(10):3881--3899, October 2014.
\newblock Publisher: American Meteorological Society Section: Monthly Weather Review.

\bibitem{schultz_can_2021}
M.~G. Schultz, C.~Betancourt, B.~Gong, F.~Kleinert, M.~Langguth, L.~H. Leufen, A.~Mozaffari, and S.~Stadtler.
\newblock Can deep learning beat numerical weather prediction?
\newblock {\em Philosophical Transactions of the Royal Society A: Mathematical, Physical and Engineering Sciences}, 379(2194):20200097, February 2021.
\newblock Publisher: Royal Society.

\bibitem{shi2015convolutional}
Xingjian Shi, Zhourong Chen, Hao Wang, Dit-Yan Yeung, Wai kin Wong, and Wang chun Woo.
\newblock Convolutional lstm network: A machine learning approach for precipitation nowcasting, 2015.

\bibitem{pytorchvgg}
Karen Simonyan and Andrew Zisserman.
\newblock Very deep convolutional networks for large-scale image recognition, 2015.

\bibitem{tabata_recalibration_2019}
Tasuku Tabata, Viju~O. John, Rob~A. Roebeling, Tim Hewison, and Jörg Schulz.
\newblock Recalibration of over 35 {Years} of {Infrared} and {Water} {Vapor} {Channel} {Radiances} of the {JMA} {Geostationary} {Satellites}.
\newblock {\em Remote Sensing}, 11(10):1189, January 2019.
\newblock Number: 10 Publisher: Multidisciplinary Digital Publishing Institute.

\bibitem{tian_tropical_2019}
Wei Tian, Wei Huangwei, Xiaolong Xu, and Chao Wang.
\newblock Tropical {Cyclone} {Maximum} {Wind} {Estimation} from {Infrared} {Satellite} {Data} with {Integrated} {Convolutional} {Neural} {Networks}.
\newblock In {\em 2019 {International} {Conference} on {Internet} of {Things} ({iThings}) and {IEEE} {Green} {Computing} and {Communications} ({GreenCom}) and {IEEE} {Cyber}, {Physical} and {Social} {Computing} ({CPSCom}) and {IEEE} {Smart} {Data} ({SmartData})}, pages 575--580, July 2019.

\bibitem{velden_dvorak_2006}
Christopher Velden, Bruce Harper, Frank Wells, John~L. Beven, Ray Zehr, Timothy Olander, Max Mayfield, Charles~“Chip” Guard, Mark Lander, Roger Edson, Lixion Avila, Andrew Burton, Mike Turk, Akihiro Kikuchi, Adam Christian, Philippe Caroff, and Paul McCrone.
\newblock The {Dvorak} {Tropical} {Cyclone} {Intensity} {Estimation} {Technique}: {A} {Satellite}-{Based} {Method} that {Has} {Endured} for over 30 {Years}.
\newblock {\em Bulletin of the American Meteorological Society}, 87(9):1195--1210, September 2006.

\bibitem{wang_review_2022}
Zhen Wang, Jun Zhao, Hong Huang, and Xuezhong Wang.
\newblock A {Review} on the {Application} of {Machine} {Learning} {Methods} in {Tropical} {Cyclone} {Forecasting}.
\newblock {\em Frontiers in Earth Science}, 10, 2022.

\bibitem{yamaguchi_tropical_2018}
Munehiko Yamaguchi, Hiromi Owada, Udai Shimada, Masahiro Sawada, Takeshi Iriguchi, Kate~D. Musgrave, and Mark DeMaria.
\newblock Tropical {Cyclone} {Intensity} {Prediction} in the {Western} {North} {Pacific} {Basin} {Using} {SHIPS} and {JMA}/{GSM}.
\newblock {\em Sola}, 14:138--143, 2018.

\end{thebibliography}
    
\newpage

\appendix 

\section{Additional Benchmark Results}

We already introduced benchmark results for three types of tasks in Section~\ref{sec-benchmarks}. The following are additional results of the same tasks, which were omitted due to space limitations. 

\subsection{Analysis for the Intensity}

\begin{table}[t]
    \caption{Results of the classification task for three architectures and three types of input.}
    \label{tab:classification-grade}
    \centering
    \begin{tabular}{ccccc} 
    \toprule
    \textbf{Acc (\%)} & Full ($512\times512$) & Resized ($224\times 224$) & Cropped ($224\times 224$)
    \\  \midrule
    VGG & 70.0 ($\pm$1.4) & 67.2 ($\pm$0.2) & 69.0 ($\pm$0.6)
    \\  \midrule
    ResNet18 & 66.6 ($\pm$0.7) & 66.3 ($\pm0.9$) & 67.8 ($\pm0.2$)
    \\  \midrule
    ViT & / & 62.0 ($\pm$0.2) & 64.9 ($\pm$1.2)
    \\ 
               \bottomrule
        \end{tabular}

\end{table}

In comparison to the regression task in Section~\ref{sec-benchmarks}, we summarize the result of the classification task. Here the target value is the categorical value of grades 2, 3, 4, 5, and 6. To compare the performance of the three types of architectures, namely VGG, ResNet18, and Vision Transformer (ViT), we run 2 training sessions on three types of input images. 
For all experiments, we used a learning rate of $10^{-4}$, batch size of 16, an 80/20 train/test split by sequence, the SGD optimizer, and 50 epochs. ViT for the full-size images was not performed due to the lack of memory.  

Table~\ref{tab:classification-grade} summarizes the result of classification in accuracy. VGG is slightly better than ResNet18, while ViT performs the worst. This result indicates that the dataset size may not be large enough for models that performed worse. However, we need to work in two directions to explore better models. First, we need to explore deeper architectures, such as ResNet50 or even deeper ResNet, to see how the depth of the network relates to the performance. Second, we need to increase the number of epochs for some models, because the best accuracy of ViT was reached during the last epochs. 

Note that the result of Table~\ref{tab:classification-grade}, showing a classification accuracy of 70\% at best, is likely to be an underestimate of the actual performance. This is due to the definition of grades, used as the target values of the classification task. As stated earlier in Section~\ref{sec-analysis}, grades 3, 4, and 5 denote a tropical cyclone with the ascending order of intensity, while grades 2 and 6 denote a tropical depression and an extra-tropical cyclone respectively. These categorical values, however, are continuous and ambiguous as labels for classification. 

First, grades 3, 4, and 5 are continuous categories and hence the boundary between categories is not clear-cut. Considering the potential annotation errors in the best track dataset as addressed in Section~\ref{sec:track}, misclassification between neighboring categories should not be considered fatal mistakes. Second, grades 2 and 3 are also neighboring categories that differ only in intensity, whereas grades 3-5 and 6 differ in the structure of the cyclone. This suggests that the categories have semantics that may have different impacts on misclassification. Hence the introduction of a more realistic loss function between categorical transitions is expected to alleviate these problems. 

Considering the difficulty in evaluating the classification task, we suggest that the regression task, especially on the central pressure, has a less ambiguous definition of the task. This is the reason that we introduced the result of the regression task in the main text. 

\subsection{Reanalysis for the Intensity}

In Section~\ref{sec-reanalysis}, the reanalysis task was performed on data splits by satellite generations. To observe if the dataset size has a valuable positive impact on the model performance, instead of sampling 208 sequences from each generation, we used the entire generation's sequences as the dataset to train each model (resulting in a train set size of 531 sequences, 183 sequences, and 167 sequences respectively). We then applied the same methodology used in Section~\ref{sec-reanalysis} to train 3 ResNet18 models for 101 epochs with 6 training sessions.

Comparison of Table~\ref{tab:additional-reanalysis-table} with Table~\ref{tab:reanalysis-table} suggests that a larger dataset size may outweigh any negative impact of inter-generational bias, as the model trained on the first bucket (the largest) performed the best. 

\begin{table}[t]
    \caption{The results of regression tasks for each satellite generation using the entire generation (values in hPa).}  
     \label{tab:additional-reanalysis-table}
    \centering
    \begin{tabular}{ccccc} 
    \toprule
    \textbf{RMSE} & Train the First & Train the Second & Train the Third
    \\  \midrule
    Test the First & 9.16 ($\pm$0.05) & 10.43 ($\pm$0.09) & 10.51 ($\pm$0.11)
    \\  \midrule
    Test the Second & 9.23 ($\pm$0.09) & 10.39 ($\pm$0.18) & 10.16 ($\pm$0.09)
    \\  \midrule
    Test the Third & 9.56 ($\pm$0.11) & 9.96 ($\pm$0.10) & 10.22 ($\pm$0.08)
    \\     \bottomrule
        \end{tabular}
\end{table}

\section{Data Collection}

\subsection{Image Dataset}

\paragraph{Accessibility}
The satellite image dataset, as introduced in Section~\ref{sec:image}, was built on geostationary meteorological satellites, Himawari-1 to Himawari-9 from JMA and GOES-9 from NOAA. The data was delivered through the following organizations.
\begin{enumerate}
    \item Japan Meteorological Business Support Center (1978-1995, 2003-2022)
\item Institute of Industrial Science, The University of Tokyo (1995-2003)
\end{enumerate}

Himawari satellite data is, in principle, not copyrighted, but the current situation is far from open and easy-to-download data. Some recent data is available online, but historical satellite data, especially satellite data in the original format, becomes less accessible as time goes back. 

\paragraph{Data format}
After obtaining the original satellite data, their processing poses another challenge due to special data formats. Himawari satellite data has three types of data formats. 

\begin{enumerate}
    \item VISSR / S-VISSR format for the first-generation satellite (Himawari 1-5, GOES 9)
    \item HRIT format for the second-generation satellite (Himawari 6-7)
    \item Himawari Standard format for the third-generation satellite (Himawari 8-9)
\end{enumerate}

Due to the lack of standard open-source tools for parsing data formats of all generations, it is not easy for a user to create a data parser for the original satellite data. To solve this problem, we implemented our parser from scratch based on the official documentation published by JMA. We then converted the original data to a standard format for which users can find an open-source data loader library.  

\paragraph{Channels}
Himawari satellite data has a few channels to observe the atmosphere in different wavelengths. The visible channel ($0.6\mu$m) measures the reflection and scattering of the sunlight from the Earth, so it has advantages in observing the detailed structure of the cloud patterns. It has a higher resolution but can be used only in the daytime. The infrared channel ($11\mu$m) measures the radiation of the Earth, roughly corresponding to the temperature of the atmosphere. It has a lower resolution but can be used both day and night. Due to the continuity of the observation, we used the infrared channel to create the standard hourly satellite image dataset for typhoon images. 

Note that recent satellites have more channels. Since Himawari-5, we have the water vapor (WV) channel ($6.7\mu$m) to measure the moisture in the atmosphere, and the near-infrared (NIR) channel ($3.9\mu$m) to measure the vegetation on the ground. Images of those channels are processed and provided from the Digital Typhoon website and may be included in the future release of the Digital Typhoon dataset. 

\subsection{Track Dataset}

The track dataset, as introduced in Section~\ref{sec:track}, is based on the best track data from JMA, available at \url{https://www.jma.go.jp/jma/jma-eng/jma-center/rsmc-hp-pub-eg/besttrack.html}. We incorporated most of the best track data, such as grade, location, central pressure, maximum sustained wind, wind circles for the storm wind (50kt) and gale wind (30kt), and the indicator of landfall or passage. The best track record is available every six hours or at shorter intervals in special cases. 

Because the satellite image dataset is an hourly dataset, the best track should be interpolated for missing hours. We designed the interpolation method for each variable so that the interpolation aligns with the design of machine learning tasks that use some of those values as 'ground truth' for supervised learning. 

\paragraph{Location}
Location is interpolated using cubic splines \cite{nii1} to create a smooth trajectory connecting best track locations. 

\paragraph{Grade}
The grade is persistent throughout the interpolation interval because it is a categorical value. 

\paragraph{Central pressure}
Central pressure is interpolated using a linear function connecting best track pressures. This is a choice to conserve maxima and minima and avoid the shooting of pressures beyond observed values. At the same time, a linear function can reflect continuous changes in the intensity of the typhoon.

\paragraph{Maximum sustained wind}
The maximum sustained wind is persistent throughout the interpolation interval. The wind is a numerical value, so it matches well with the linear interpolation. However, we should consider the fact that the wind is only available for grades 3, 4, and 5, but not for grades 2 and 6, where the wind value is set to zero. Grade 2 is called a tropical depression which has a smaller maximum sustained wind than 35 knots. Because the purpose of the JMA best track is to create the record of typhoons with more than 35 knots, JMA simply does not estimate the maximum sustained wind of tropical depressions. In a similar manner, Grade 6 is called an extra-tropical cyclone, which has a different atmospheric structure from a tropical cyclone. Although the maximum sustained wind of extra-tropical cyclones may exceed 35 knots somewhere far from the center, the JMA best track does not contain the maximum sustained wind for Grade 6 for the same reason as Grade 2. 

This leads to discontinuity of the wind values at the time of formation (from grade 2 to grade 3) and transition (from grade 3-5 to grade 6) of typhoons. To avoid inappropriate linear interpolation at discontinuous points, we decided to keep the value persistent. The same rule applies to wind circles. 

\section{Data Processing Workflow}

After data collection, the original satellite image is fed into a data processing pipeline, which has been developed by the author (Asanobu Kitamoto) since 1992. The pipeline consists of software codes written in C and Perl. They are not open-sourced due to the complex structure of software. Note that an open-source software library \texttt{pyphoon2} has been released as introduced in Section~\ref{sec-benchmarks} to work with the Digital Typhoon dataset for machine learning tasks. 

The summary of the pipeline is as follows.

\begin{enumerate}
\item Parse the original satellite image and create a map-projected 2D array image with digital count as pixel values (Section~\ref{subsec-map-projection}). 
\item For each pixel value, convert digital count (integer value) to brightness temperature (floating-point value). The conversion is done in two steps (Section~\ref{subsec-calibration}).
\begin{enumerate}
\item The first step is to convert the digital count to brightness temperature using the conversion table for each sensor. 
\item The second step is to convert brightness temperature to calibrated brightness temperature using inter-calibration parameters from the recalibration project \cite{tabata_recalibration_2019,john_methods_2019}. 
\end{enumerate}
\item The calibrated brightness temperature, including a masked pixel value, is saved in the HDF5 file (Section~\ref{subsec-mask}). 
\end{enumerate}

In the following, we summarize the details of each step. 

\subsection{Map Projection}
\label{subsec-map-projection}

The map projection is used to create a typhoon-centered image. When a typhoon is observed from a satellite, the shape of the typhoon is usually distorted from the satellite's viewpoint due to the curvature of the Earth's 3D surface. Here the role of map projection is two-fold. First, the map projection creates a typhoon image where the typhoon center is located at the center of the image. Second, the map projection reduces the distortion of the typhoon by creating an image from a viewpoint above the typhoon center. 

The choice of the map projection method depends on the choice of metric properties that should be preserved at the expense of others. For the Digital Typhoon dataset, we chose the Lambert azimuthal equal-area projection to preserve the area and a circular shape around the center. This is in contrast to the choice of map projection in the HURSAT dataset (Section~\ref{sec-hursat}), where their choice, equirectangular projection (lat/long grid), does not preserve any metric properties. But at least, equirectangular projection is better than a simple image cropping from the original satellite image, because the distortion of the shape is reduced during the map projection. 

It is an interesting question whether we can choose the "best" map projection for a machine learning dataset. Due to the wide variety of map projections, evidenced by about 150 map projections implemented in the open-source library \texttt{proj} (as of October 2023), a comprehensive quantitative comparison of map projections under a metric is not feasible. Instead, we claim that our map projection satisfies the desirable characteristics of map projection for this dataset. For example, it is well known that Mercator projection significantly distorts the shape to the north, hence it is not an appropriate choice to focus on shape. On the other hand, the choice of azimuthal projections over other projections can be justified by the fact that the center of a tropical cyclone is a special point for its tracking. Moreover, among azimuthal projections, we can choose either equidistant or equal-area. We chose the equal-area because we assume the area of cloud pixels is a more important indicator of the intensity of tropical cyclones than the distance from the center. 

Next, we consider the size of the typhoon image after map projection. Here, the size has two aspects, either the size of the scene or the size of the image. First, about the size of the scene on the original satellite image, we refer to the maximum circle of gale wind (30kt), whose diameter is 1275nm (nautical mile) or 2361km recorded by Typhoon 199713. Based on this record, we set the diameter of the scene as 2500km. Second, about the size of the image, we decided to use a square image of 512 pixels. This amounts to a resolution of 4.88km/pixel, which is close to the maximum resolution of 5km for the infrared image of old-generation satellites. 

The last choice is the geometric transformation for map projection or output-to-input mapping. Some typical choices are bi-linear or bi-cubic interpolation, but they could introduce non-existent pixel values due to the mixture of observation values from multiple pixels, and conversion from digital count to brightness temperature, which will be addressed below, becomes more complicated. To avoid this problem, we chose a simpler nearest-neighbor method that preserves original pixel values. Note that some pixel values are copied more than once when the resolution of the original satellite image is not enough due to the curvature of the Earth. 

\subsection{Calibration}
\label{subsec-calibration}

The Digital Typhoon dataset is the collection of 40+ years of satellite data from 10 different satellites, and calibration for removing biases in each sensor is required to create a homogeneous long-term dataset. The original satellite image is recorded as the collection of the digital count, which is an integer value that records the response of the sensor. Depending on the precision of the sensor, the digital count is represented by 6 bits for a low-precision sensor and 12 bits for a high-precision sensor. From these digital counts before calibration, the following procedure is designed to produce calibrated values. 

The first step is to convert the digital count to brightness temperature, which is a physical value measured in Kelvin (K) representing the temperature of the Earth such as cloud top, ground, and ocean surface. The conversion table from the digital count to the brightness temperature is provided for each sensor based on the initial calibration of the sensor. 

The second step is to apply a recalibration equation to convert brightness temperature to calibrated brightness temperature. This recalibration method is effective for inter-calibration across satellite sensors to obtain even better homogeneous long-term observation data for climate change research. This calibrated brightness temperature is included in the Digital Typhoon dataset.

Note that some machine-learning papers use typhoon images that are scraped from public typhoon information websites, including the Digital Typhoon website. Using scraped images for machine learning research, however, suffers from a number of problems. First, the JPEG format contaminates pixel values with lossy compression. Second, the pixel value of a popular image format, such as 0-255 in 8 bits, has been applied to an unknown scaling function and hence does not have a physical meaning. Third, some scraped images contain graphical elements such as the coastline and lat/long lines, and they introduce unnecessary noise. The Digital Typhoon dataset does not have those problems and machine learning researchers can focus on designing algorithms and models on a clean dataset.

\subsection{Masking}
\label{subsec-mask}

The final step of the pipeline is to save calibrated brightness temperature to a file. Because each pixel takes a floating point value, we decided to use the HDF5 (Hierarchical Data Format 5), which is a popular data format for scientific applications to deal with a multidimensional array of floating point values. 

Here a problem arises with pixels that do not have observation data. This happens in the following situations. 

\begin{enumerate}
\item A pixel is out-of-frame. It occurs when a typhoon is located at the periphery of a satellite image, and hence a typhoon-centered image overlaps with the boundary of the satellite image. It also occurs when a satellite observation is partial due to scheduled maintenance or emergency operations. 
\item A pixel is contaminated by noise. It occurs in old satellites when the sensor malfunctions. These noises can be easily detected when the pixel value takes the minimum or maximum values of the range but is more difficult to detect when the noise is only visible as an irregular spatial pattern. 
\end{enumerate}

In HDF5, there is no standard way to represent a pixel value without observation (invalid or null values). Hence we used a temperature of 130.0, which is below the valid brightness temperature (130K or about -140 degrees Celsius). They can be used to mask invalid pixels when machine-learning models can properly treat masked pixels. 

Note that we can further normalize the brightness temperature as input to machine learning models. In our benchmarking experiments, we used the standard normalization procedure to map a brightness temperature of $[170, 300]$ to a normalized value of $[0,1]$. In this setting, the masked pixels are mapped to 0 together with other extremely cold pixels, but note that the minimum temperature of 170K rarely happens in the natural environment. 

The final step is to decide if we put the image into the dataset. If an image has less than 30 percent of masked pixels, the image is included in the dataset with masked pixels. Otherwise, the image is not included and becomes missing data in the dataset. Hence the Digital Typhoon dataset consists of both perfect images without noise and imperfect images with noise or occlusion. To deal with imperfect images, \texttt{pyphoon2} has a filtering option to exclude some images from the machine-learning task according to the percentage of masked pixels, and it is the user's responsibility to treat filtering appropriately for their machine-learning tasks.

\section{Dataset Organization and Updating}

\begin{figure}
    \centering
    \includegraphics[width=1.0\textwidth]{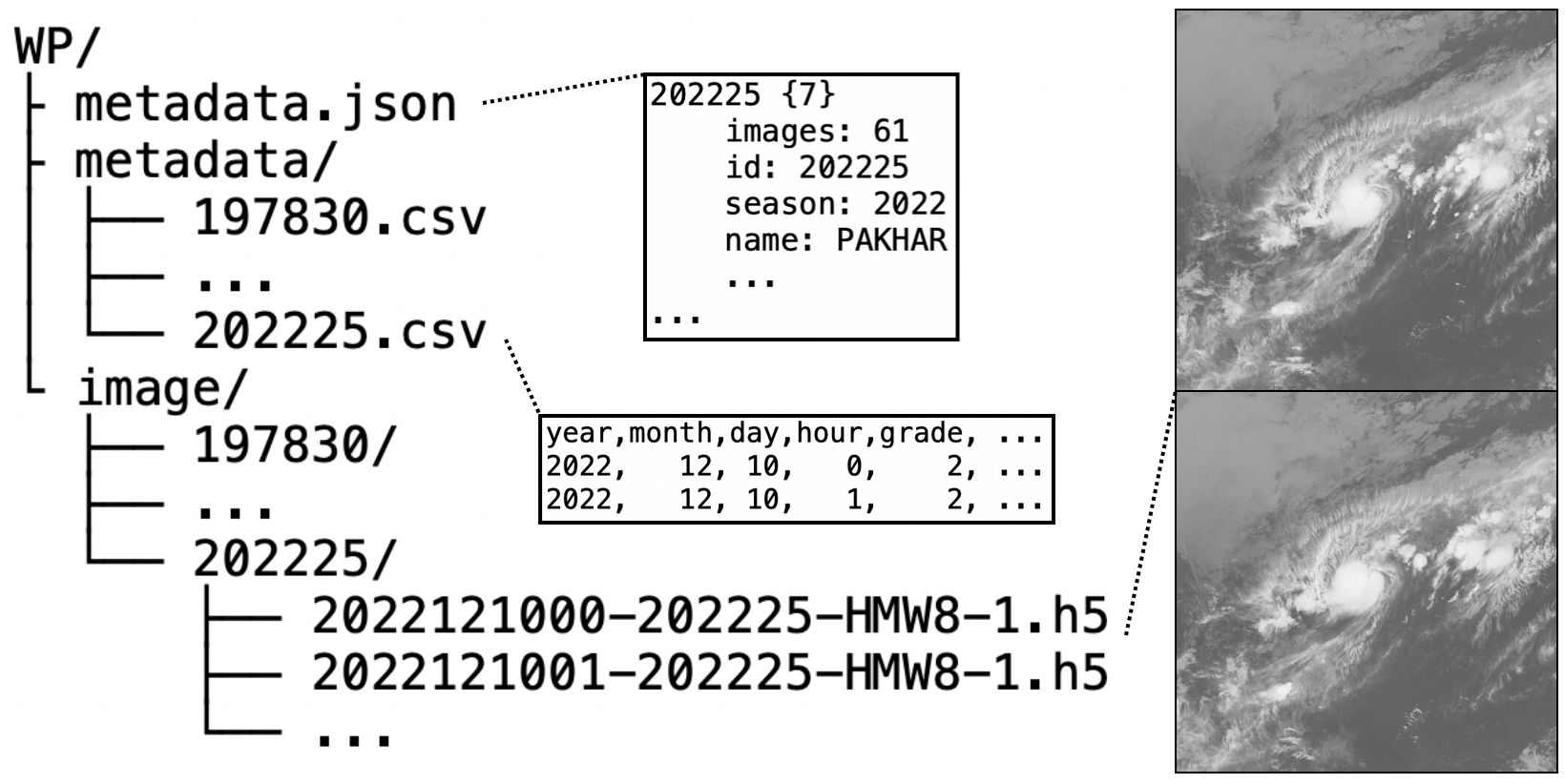}
    \caption[width=0.8\textwidth]{Directory structure of the Digital Typhoon dataset}
    \label{fig:directory}
\end{figure}

The directory structure of the dataset is summarized in Figure~\ref{fig:directory}. The dataset consists of the image directory, containing the list of HDF5 files for hourly images grouped by each typhoon, and the metadata directory, containing one file for each typhoon with the record of best track data and the quality of the image of the corresponding observation time. At the top level, we also provide the metadata.json file, which contains information about each typhoon in the dataset, such as the season, the number of images, and the name of the typhoon.  

The Digital Typhoon dataset will be updated when all the best track for a typhoon season is available, which is typically January or February of the next year in the case of the JMA best track. At the same time, the best track for some old typhoons may be updated based on the result of recent reanalysis activities. 
Hence a new dataset will include new data for the last season in addition to data for the past seasons with minor updates. 

The current dataset includes data until the 2022 season and the whole dataset is zipped into one file with the size of 54GB. In January or February 2024, we will create a dataset until the 2023 season, and replace the current ZIP file with a new one. The old data may be kept for a while for reproducible research, but roughly speaking, this is not necessary because the research is mostly reproducible by just filtering the data for the latest season. Hence it is important to record the last season of the experiments when reporting the results of experiments.  

In the future, we may create related datasets to the main dataset. For example, we can release a temporary dataset for the current season, or even real-time datasets for experimenting with forecasting in real-time. In addition, we can release another dataset with higher spatial and temporal resolutions for third-generation satellites. They will be a part of the Digital Typhoon dataset as separate ZIP files so that users can choose the dataset for their needs. 

\section{Responsible Use}

The dataset should be used responsibly so that machine learning results do not mislead the public in terms of disaster preparedness and response. For example, in Japan, the public announcement of typhoon forecasting in real-time is only allowed for JMA, and for others, it is strictly prohibited by law. There may be similar regulations in other countries for public safety. Because typhoon forecasting is critical information for life-and-death decision-making, responsible use is required for real-time forecasting by machine learning. However, this regulation does not apply to machine learning benchmarks for historical datasets.

Another issue for responsible use is aligning with other scientific communities, such as atmosphere and climate sciences when publishing controversial results that might influence the future of human society. As discussed in Section \ref{sec:track}, long-term data has many types of data quality issues such as biases that require careful modeling to produce reliable results. Machine learning research without careful consideration of those issues may arrive at misleading conclusions that raise controversies without solid scientific evidence. This problem can be alleviated by communicating with domain experts to cross-check the results from a broader perspective. 

\section{Dataset Availability}

The Digital Typhoon dataset will be maintained by the Digital Typhoon project at the National Institute of Informatics. This project started in 1999 and has been continuously maintaining the data processing pipeline for over 20 years. This history proves that we already established a robust system for the maintenance of the dataset. 

Given that satellite images are publicly accessible observational data, we have designated the license for the Digital Typhoon dataset as the Creative Commons Attribution License (CC BY). Attribution to the dataset can be shown as follows. 

\begin{quote}
    Digital Typhoon Dataset (National Institute of Informatics) 
    doi: \url{https://doi.org/10.20783/DIAS.664}
\end{quote}

Additionally, we credit two organizations as the data source. First, JMA is the organization responsible for the official data distribution. Most of our data came from JMA through JMBSC (Japan Meteorological Business Support Center). Second, satellite data between 1995 and 2003 were downloaded from the Institute of Industrial Science (IIS), The University of Tokyo, which was in charge of satellite data receiving stations and data archiving systems. 

In addition, the following pages provide related information about the dataset. 

\paragraph{Digital Typhoon}
\url{http://agora.ex.nii.ac.jp/digital-typhoon/dataset/}

Digital Typhoon website offers a variety of services to explore a wide range of typhoon-related data. The official page of the dataset is available on this website. 

\paragraph{GitHub}
\url{https://github.com/kitamoto-lab/digital-typhoon/}

The GitHub page provides code to work with the dataset to train machine-learning models used for benchmarking. The code is provided with the MIT license.

\paragraph{Hugging Face}
\url{https://huggingface.co/kitamoto-lab}

The Hugging Face page provides model weights and some codes for using them. The model is provided with the MIT license.

\paragraph{DIAS}
\url{https://diasjp.net/}

The Digital Typhoon dataset is also available from a data repository DIAS (Data Integration and Analysis System). DIAS is a Japanese data repository for earth science and environmental datasets, hence it is a suitable place to store the dataset for long-term preservation. It also provides the dataset DOI (Digital Object Identifier) doi:10.20783/DIAS.664 or \url{https://doi.org/10.20783/DIAS.664} as a persistent identifier. 

\end{document}